\pdfoutput=1
\documentclass[11pt]{article}

\usepackage[final]{acl}
\usepackage{times}
\usepackage{latexsym}
\usepackage[T1]{fontenc}
\usepackage[utf8]{inputenc}
\usepackage{microtype}
\usepackage{inconsolata}
\usepackage{graphicx}
\usepackage{amsmath}
\usepackage{amsthm}
\usepackage{amsfonts}
\usepackage{enumitem}
\usepackage{xspace}
\usepackage{ifthen}
\usepackage{soul}

\newcommand\cites[1]{\citeauthor{#1}'s\ (\citeyear{#1})}

\title{Evaluating Spatiotemporal Consistency in Automatically Generated Sewing Instructions}

\author{Luisa Geiger\textsuperscript{*+}, Mareike Hartmann\textsuperscript{+}, Michael Sullivan\textsuperscript{+}, Alexander Koller\textsuperscript{+} \\
\textsuperscript{*}Congree Language Technologies GmbH \\
\texttt{lgeiger@congree.com} \\
\textsuperscript{+}Department of Language Science and Technology, Saarland University \\
\texttt{\{mareikeh, msullivan, koller\}@coli.uni-saarland.de}
}

\begin{document}
\maketitle
\begin{abstract}

In this paper, we propose a novel, automatic tree-based evaluation metric for LLM-generated step-by-step assembly instructions, that more accurately reflects spatiotemporal aspects of construction than traditional metrics such as BLEU and BERT similarity scores. We apply our proposed metric to the domain of sewing instructions, and show that our metric better correlates with manually-annotated error counts as well as human quality ratings, demonstrating our metric's superiority for evaluating the spatiotemporal soundness of sewing instructions. Further experiments show that our metric is more robust than traditional approaches against artificially-constructed counterfactual examples that are specifically constructed to confound metrics that rely on textual similarity. 

\end{abstract}

\section{Introduction}
\label{sec_intro}


\begin{figure}[t]
    \centering
    \includegraphics[width=74mm, height=111mm]{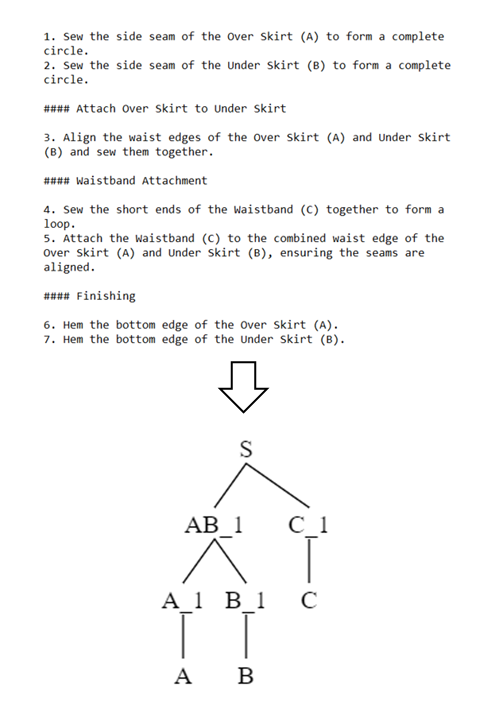}
    \caption{Illustration of our tree-based evaluation metric. Instructions are automatically converted a tree that encodes the order and components of each step. These automatically-extracted trees are then compared to gold trees to yield a score.}
    \label{fig:generaloverview}
\end{figure}


Creating consistent, high quality instructions in any domain can be a difficult process.
Much research has investigated the ability of Large Language Models (LLMs) to generate such instructions, most commonly in the domain of cooking recipes
\citep[e.g.][]{li-etal-2024-understanding-retrieval, Salvador_2019_CVPR}.

High quality instructions make it as easy as possible for the reader to connect the textual instructions to the required actions in the physical world.
For the instructions to make sense with regard to the physical world, the LLM that is generating them needs an understanding of the current world state. This necessitates spatial (where objects are located) and temporal (when objects change their location) awareness: although LLMs have continuously improved on a wide variety of benchmarks, 
they still have difficulties with consistency in the area of spatiotemporal reasoning \cite{aghzal2025largelanguagemodelsgood}.



In particular, the generation of textual sewing instructions is a multimodal reasoning problem that necessitates spatiotemporal awareness of multiple objects and their surroundings, as well as knowledge of the current world state: specifically, the current assembly state of the garment at each step of the instructions. 

This task is similar to other instruction generation tasks such as cooking recipe generation and Minecraft instruction generation \citep[e.g.][]{narayan-chen-etal-2019-collaborative}, in that the generation task consists of generating procedural textual instructions, including explicit or implicit representations/descriptions of the current world state.

However, generating sewing instructions differs from these more widely-researched tasks, due to the more complex operations that need to be performed.

The complexity of this process lends to sensitivity to even slight errors: incorrect use of terminology or ambiguous explanations of assembly operations can lead to catastrophic misunderstandings between writer and the reader. While this can also be the case for other instruction generation tasks, the increased complexity involved in the physical manipulation of fabric pieces leads to a far more spatiotemporally sensitive process than (for example) stacking blocks.



Since the quality of sewing instructions heavily depends on their ability to capture the correct way to assemble the pieces, evaluating generated sewing instructions in meaningful way is not a trivial task.
Common evaluation metrics such as BLEU \cite{papineni-etal-2002-bleu} and BERT-Score \cite{zhang2019bertscore} only focus on superficial similarities between the generated instructions and a given set of gold reference instructions (see Sections \ref{sec_relwork} and \ref{sec_experiments}).

In this paper, we introduce an automatic, tree-based evaluation metric (Section \ref{sec_tree_eval}), that mitigates the shortcomings of traditional evaluation metrics with respect to the task of instruction generation. We compare our evaluation metric to BLEU, ROUGE-L \cite{lin-2004-rouge}, and BERT-Score on sewing instructions generated via a range of prompting strategies, demonstrating the versatility and robustness of our approach with respect to varying methods of instruction generation (Section \ref{sec_experiments}). 

We show that our metric better correlates with manually-annotated error counts in the model-generated instructions, indicating that our approach better reflects spatiotemporal correctness than competing metrics. 

To further highlight the sensitivity of our approach to correctness, we construct an artificial dataset in which the steps of the model-generated instructions are randomly permuted: while our metric is highly sensitive to these nonsensical instructions, the traditional metrics fail to meaningfully reflect any difference between the original and permuted instructions. 

Additionally, we find that our metric is weakly positively correlated with subjective human ratings, while all other, traditional similarity measures are negatively correlated.







All prompts, generated instructions, input images, code files, and data required to replicate these experiments are available on GitHub\footnote{\url{https://github.com/coli-saar/generatingsewinginstructions}}.

\section{Related Work}
\label{sec_relwork}

\paragraph{Multimodal Reasoning.}

\citet{liu2024visualwebbenchfarmultimodalllms} evaluate a range of Multimodal LLMs (MLLMs) across a variety of web tasks, and find that many models do not exceed random-chance performance on action prediction and grounding tasks, speaking to these models' limited reasoning and grounding abilities\textemdash both critical skills for sewing instruction generation (as discussed in Section \ref{sec_intro}). Although GPT-4V \cite{achiam2023gpt} and Claude \cite{anthropic2024claude} outperform their open-source counterparts by a notable margin, GPT-4V\textemdash the best-performing model they evaluate\textemdash only achieves an average score of 64.6 out of 100, indicating that there is still much room for improvement, even among closed-source MLLMs.

\citet{ji2022abstractvisualreasoningtangram} investigate human and LLM abtract visual reasoning abilities through the use of tangram puzzles \cite{hawkins2020characterizing}: both models that they evaluate\textemdash CLIP \cite{pmlr-v139-radford21a} and ViLT \cite{kim2021viltvisionandlanguagetransformerconvolution}\textemdash demonstrate limited abstract reasoning abilities in comparison to human participants. However, \citet{ji2022abstractvisualreasoningtangram} find that explicit descriptions and color-coded highlights of various tangram components improve performance for both humans and MLLMs, suggesting that similar knowledge-augmentation can be leveraged to improve model-generated sewing instructions.  

\paragraph{Textual Spatiotemporal Reasoning and Planning.}

Although the findings of \citet{lyu-etal-2020-reasoning} indicate that fine-tuning and careful data-engineering can improve the common-sense reasoning abilities of then-SoTA LMs\textemdash i.e.\hspace{1mm}BERT \cite{devlin-etal-2019-bert}, XLNet \cite{yang2019advances}, RoBERTa \cite{liu2019robertarobustlyoptimizedbert}, and GPT-2 \cite{Radford2019LanguageMA}\textemdash to near-human performance, \citet{aghzal2025largelanguagemodelsgood} suggest that the capabilities of LLMs are limited when it comes to long-term planning and spatial reasoning. Specifically, \citet{aghzal2025largelanguagemodelsgood} find that in-context learning (ICL) does not help GPT-4 avoid obstacles\textemdash although ICL does improve the model's ability to reach the goal\textemdash in textual grid-navigation environments. While chain-of-thought 
\citep[CoT;][]{wei2022chainofthought} prompting does improve obstacle-avoidance abilities, these abilities still degrade as obstacle number and distance to the goal increase. 

Of particular concern to the topic at hand are \cites{wu-etal-2024-reasoning} findings that LLM's reasoning performance degrades substantially when they are presented with counterfactual instances: reasoning problems in which well-known rules are altered, such as chess puzzles in which the piece types' valid moves are modified. This indicates that LLMs may struggle to accomplish a task such as sewing-instruction generation\textemdash examples of which are unlikely to occur frequently in the models' training data.

\paragraph{Cooking Recipe Generation.}

Given the similarities between the two tasks, it stands to reason that many findings in the area of cooking-recipe generation\textemdash a far more well-researched task\textemdash should translate to sewing-instruction generation.

\citet{Salvador_2019_CVPR} propose an approach for generating cooking recipes under which the model first predicts ingredients before generating a recipe. The authors find that their method generates higher quality recipes\textemdash and improves in ingredient prediction\textemdash over previous baselines, and creates more compellingly written recipes than retrieval methods, according to human judgment. Similarly, \cite{chandu-etal-2019-storyboarding} improve over baseline cooking-generation approaches via a storyboarding approach that imposes hierarchical structure on the model's step-by-step instruction-generation reasoning process. 

On the other hand, \citet{liu2024retrievalaugmentedrecipegeneration} use Retrieval Augmented Generation (RAG) to address the common problem of hallucination in cooking-recipe generation with text generation.

\paragraph{Evaluation Metrics.}

Evaluation metrics for LLM-generated text broadly fall into one of three categories \cite{celikyilmaz2021evaluationtextgenerationsurvey}: (i) human-centric, which focus on manual evaluation of generated text; (ii) untrained and automatic, which typically compute the similarity between the generated text and a reference text \citep[e.g.][]{papineni-etal-2002-bleu}; and (iii) machine-learned, which also compute the similarity between the generated text and a reference text, but via a neural model \citep[e.g.][]{zhang2019bertscore}.

Although\textemdash by definition\textemdash it most accurately reflects human judgment, manual text evaluation is very time-consuming and expensive, hindering its employment at scale. On the other hand, automatic evaluation metrics rely purely on surface-level similarity with the reference text, which is not always applicable for instruction-generation tasks: for example, task-oriented text generation allows for a large degree of diversity in the generated texts, in which case the usefulness of such similarity metrics is limited.

Furthermore, \citet{ostmeier-etal-2024-green} note that common textual-similarity evaluation metrics such as BLEU \cite{papineni-etal-2002-bleu} and ROUGE \cite{lin-2004-rouge} lack the ability to measure factual correctness, which is critical for evaluating a wide variety of tasks. 

To address this deficiency in the medical domain, the authors introduce the GREEN metric, a quantitative metric derived from LLM-as-judge evaluation \citep[][use GPT-4]{ostmeier-etal-2024-green}. Although they find that GREEN scores are highly correlated with human expert evaluations for radiology reports, the critical weakness of this approach lies in its reliance on an LLM to evaluate factual correctness: in a domain where the model lacks factual knowledge\textemdash such as sewing-instruction generation\textemdash it is not feasible to rely on an LLM as a critical component of the evaluation metric.


\section{Task and Dataset}
\label{sec_datasetandbackground}

In this section, we provide a brief overview of sewing patterns in general (Section \ref{sec_sewingpatterns}), and the particular sewing-pattern dataset that we employ in this work (Section \ref{sec_dataset}).

\subsection{Sewing Patterns}
\label{sec_sewingpatterns}

\begin{figure}[t]
\centering
\includegraphics[width=77mm, height=43mm]{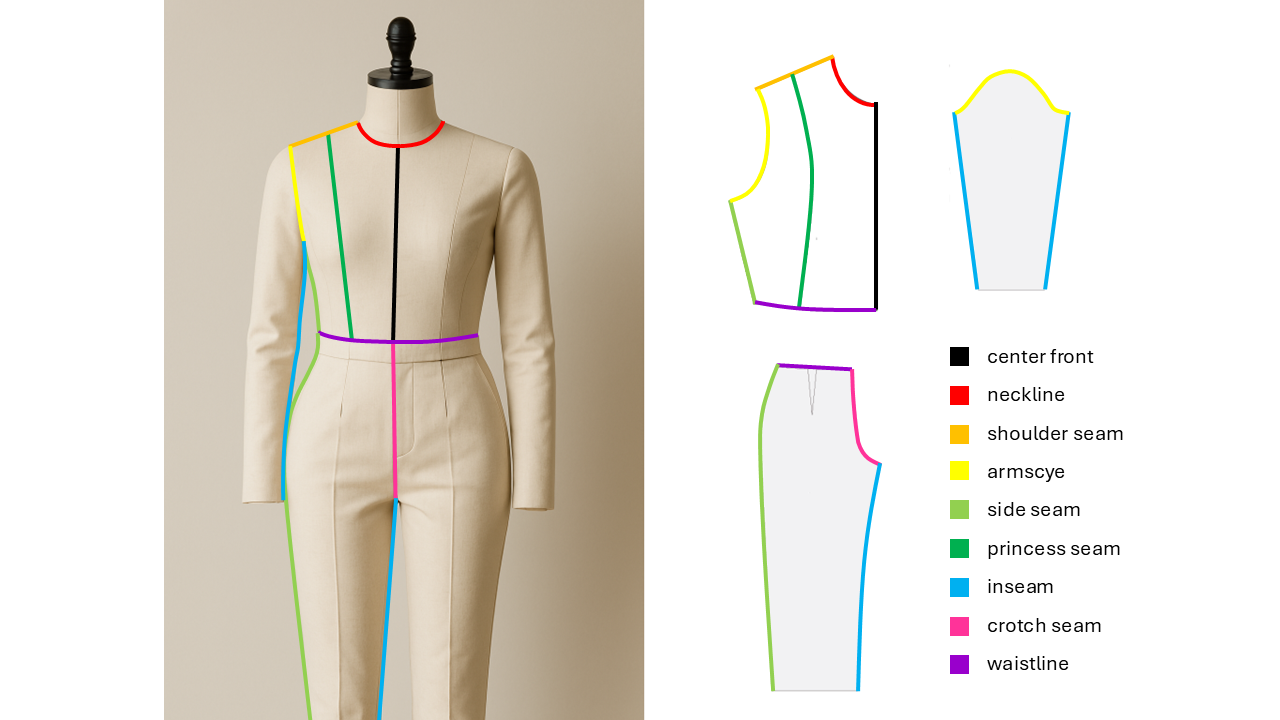}
\caption{Example of a sewing pattern (right; edges colored in this figure to illustrate attachment location) and a mannequin wearing the completed garment built from the pattern (left). (Left-hand image generated by ChatGPT; front bodice pattern from \url{https://www.moodfabrics.com/blog/}; sleeve and front pant leg patterns from \url{https://sewguide.com/princess-seams/})}
\label{fig:howsewingpatternswork}
\end{figure}

Sewing patterns are templates used to guide the construction of a garment while sewing. These paper patterns are placed on top of fabric, in order to cut out the individual pieces: the fabric pieces are then assembled into a garment (see Figure \ref{fig:howsewingpatternswork}).

Since garments are typically symmetrical, many patterns are made to show only the left or the right half\textemdash for example, Figure \ref{fig:howsewingpatternswork} shows pattern pieces for the left half of the bodice, the left sleeve, and the left front pant leg.
To create a symmetrical garment, those pieces are either cut multiple times or ``on the fold'': i.e.\hspace{1mm}by folding the fabric in half and placing the paper pattern piece near the folded edge so that the pattern cutter cuts through two layers of fabric, resulting in a perfectly symmetrical piece when it is unfolded (see Figure \ref{fig:cuttingonthefold} in the Appendix). For a bodice piece such as that in Figure \ref{fig:howsewingpatternswork}, it is common to place the center front on the folded edge of the fabric, so that the center front acts as the mirror axis and the resulting piece covers both the left and the right front side of the torso.



\subsection{Dataset}
\label{sec_dataset}
\begin{figure}
    \centering
    \includegraphics[width=\columnwidth]{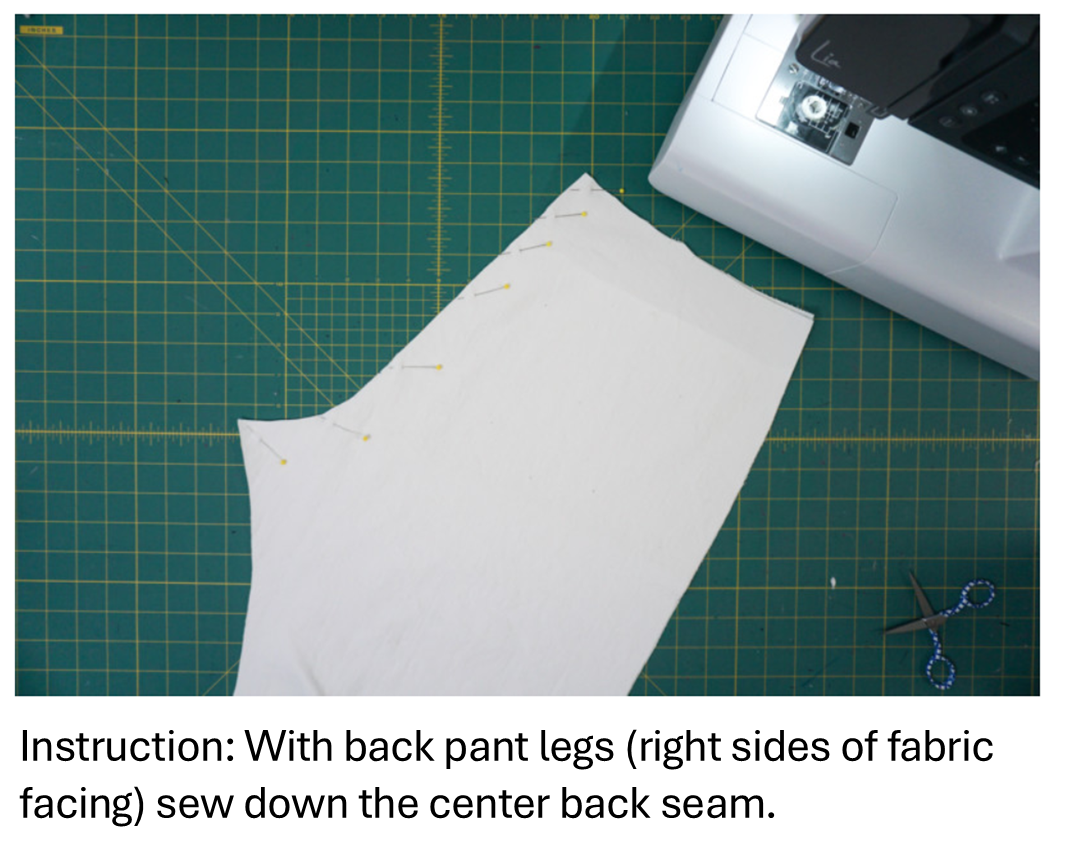}
    \caption{Example of sewing instructions in our dataset.
    }
    \label{fig:websitebeispiel}
\end{figure}

The website moodfabrics.com\footnote{\url{https://www.moodfabrics.com/blog/}} features free sewing patterns and instructions.
The instructions for each pattern describe the different steps required to go from the pattern pieces to the finished garment.
Each step is illustrated with a picture of the current state of the process (see Figure \ref{fig:websitebeispiel}).

\begin{figure}
    \centering
    \includegraphics[width=\columnwidth]{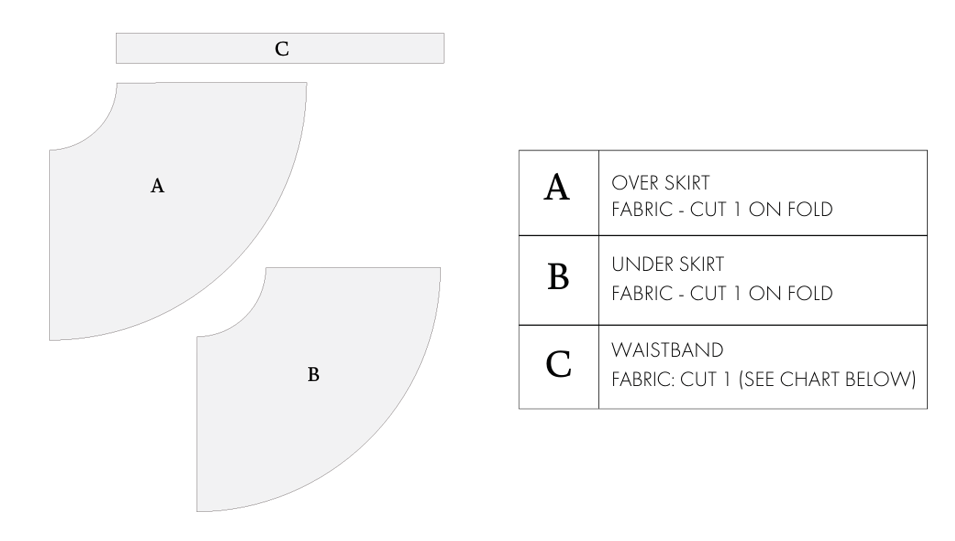}
    \caption{Example of a sewing pattern (left), with corresponding descriptions of each of its component pieces (right).}
    \label{fig:overviewanddescrMDF359}
\end{figure}

To construct our dataset, we sampled 22 patterns spanning five garment types: five shirts, five dresses, five pants, five skirts, and two jumpsuits.
Each of these patterns contains an overview picture and a description of each piece (see Figure \ref{fig:overviewanddescrMDF359}).

The pattern pieces in the overview are labeled from A to Z, allowing one to refer to the individual pieces when writing instructions.
 
In order to obtain symmetrical pieces for the left and right half of the body, pieces on the overview image have to be cut out of fabric twice (as discussed in Section \ref{sec_sewingpatterns}).
In such cases, the pattern piece description gives the instructions to cut a ``mirrored pair'' of the piece X:
the pattern cutter cuts a left (Xl) and a right (Xr) piece out of the fabric.
When the pattern piece descriptions mention cutting multiples of a piece without mirroring them, we refer to these pieces as X1, X2, X3, etc.

\section{Tree-Based Evaluation Metric}
\label{sec_tree_eval}

As discussed in Sections \ref{sec_intro} and \ref{sec_relwork}\textemdash and further demonstrated below in Section \ref{sec_experiments}\textemdash traditional evaluation metrics used for text generation are poor measures of spatiotemporal consistency. To remedy this deficiency, we introduce a tree-based evaluation metric that is designed to reflect the order of and constituent pieces involved in assembly steps. 

We first annotate each pattern with sets of gold trees (see e.g.\hspace{1mm}Figure \ref{fig:skirtassembly}), each of which encodes a valid assembly order for the garment in question (see Section \ref{sec_tree_eval_sub_annotation}). We then employ an automatic tree-extraction procedure over the model-generated instructions, which constructs trees encoding the order in which the pieces of the garment are assembled together in the instructions (see Section \ref{sec_tree_eval_sub_postprocessing}). These automatically-extracted trees are compared to their corresponding patterns' gold trees to yield the tree score (see Section \ref{sec_tree_eval_sub_score}), a measure of the spatiotemporal consistency of the generated instructions. 

\subsection{Gold Instruction Annotation}
\label{sec_tree_eval_sub_annotation}

\begin{table}[t]
    \centering
    \scalebox{0.77}{\begin{tabular}{l|c|c}
         Garment Type & Annotated Patterns & Mean Trees per Pattern \\
         \hline
         \hline
         Shirt/Top & 5 & 5,017.2 \\
         \hline
         Dress & 5 & 417.2\\
         \hline
         Pants & 5 & 1,685.2\\
         \hline
         Skirt & 5 & 3.6\\
         \hline
         Jumpsuit & 2 & 991.0\\
         \hline
         \textbf{Total} & \textbf{22} & \textbf{1,709.0}\\
    \end{tabular}}
    \caption{Overview of the annotated patterns in our dataset.}
    \label{tab:annotatedpatterns}
\end{table}

As discussed above, our tree-based evaluation metric requires the manual annotation of gold trees for each pattern in the dataset. As each garment permits multiple possible assembly orders to arrive at the same end product (see Figure \ref{fig:shirtassembly} in the Appendix), each pattern is annotated with a set of gold trees, each of which corresponds to a valid order of assembly for the garment.

There is a large number of gold trees required for each pattern (see Table \ref{tab:annotatedpatterns}): in practice, we annotate each pattern with a set of Context-Free Grammar (CFG) rules\textemdash which then generate the set of gold trees for the pattern\textemdash in order to semi-automate the annotation process.

\subsubsection{Gold Tree Structure}
\label{sec_tree_eval_sub_annotation_sub_gold}

\begin{figure*}[t]
    \centering
    \includegraphics[width=\textwidth]{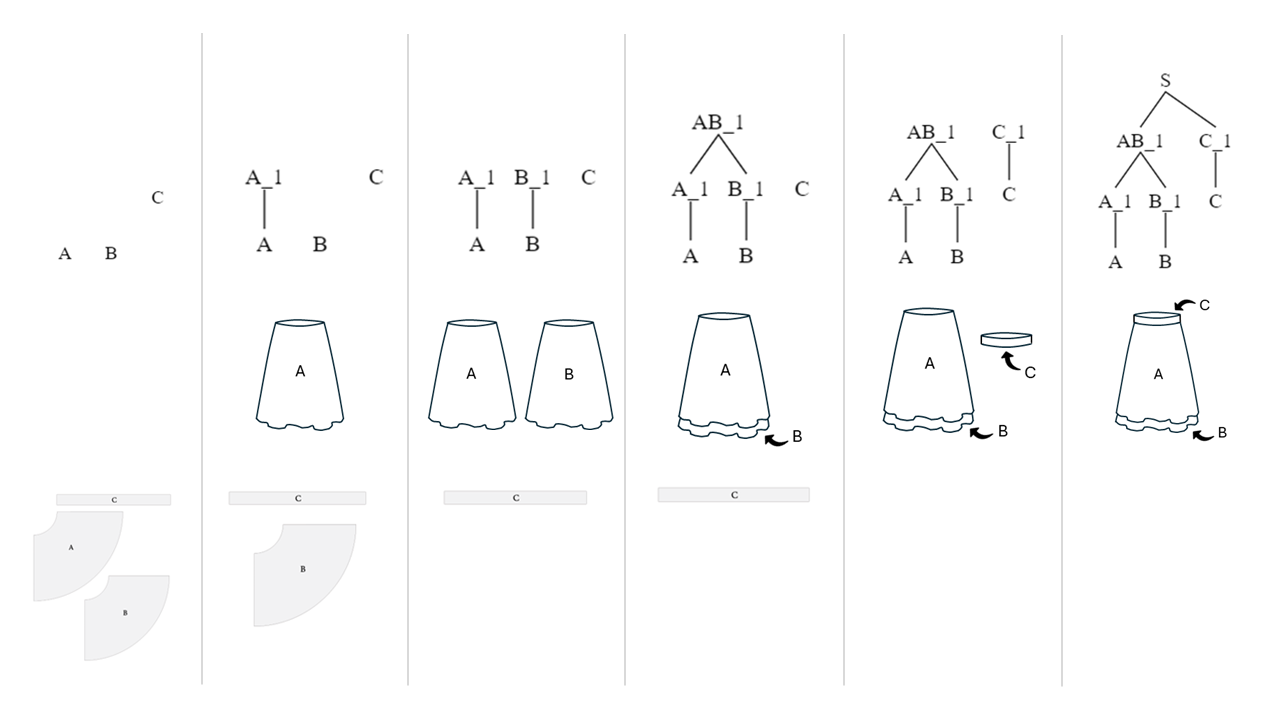}
    \caption{Step-by-step illustration of the assembly of a skirt (middle), and the corresponding tree (top). The bottom row indicates the remaining pattern pieces at each step.}
    \label{fig:skirtassembly}
\end{figure*}

For each valid assembly order, we construct a corresponding tree $t$ (see Figure \ref{fig:skirtassembly}). Each constituent piece in the garment is represented by a leaf node in $t$, and each intermediate component is represented by a node higher in the tree. For each assembly step that joins two pieces $\alpha$ and $\beta$, there is a node $\alpha\beta$ and edges $\alpha\beta\rightarrow\alpha$, $\alpha\beta\rightarrow\beta$ in $t$: the completed garment is represented by the root node $S$.

Each node in each gold tree is only permitted at most two daughter nodes. This restriction to unary- (see Section \ref{sec_tree_eval_sub_annotation_sub_self}) and binary-branching trees is imposed to facilitate the predicted tree extraction procedure (see Section \ref{sec_tree_eval_sub_postprocessing}): we impose this constraint on the gold trees to permit one-to-one comparison with the predicted trees.

We further require that the constituent piece names within the node names be in alphabetical order with respect to the root node labels, regardless of the daughter node to which each root node belongs: for example, $\text{A}\leftarrow\text{AB}\rightarrow\text{B}$ and $\text{AC}\leftarrow\text{ABCD}\rightarrow\text{BD}$ are valid subtrees, while $\text{A}\leftarrow\text{BA}\rightarrow\text{B}$ and $\text{AC}\leftarrow\text{ACBD}\rightarrow\text{BD}$ 
are not. 

Lowercase labels ``l'' and ``r'' indicating directionality (see Section \ref{sec_datasetandbackground}) stay with the label that they modify, and only serve as tie-breakers to decide the ordering between two labels with the same capital letter piece label. For example, $\text{ABl}\leftarrow\text{ABlBr}\rightarrow\text{Br}$ is a valid subtree, while $\text{ABl}\leftarrow\text{ABBlr}\rightarrow\text{Br}$ is not.

These constraints are again intended to permit one-to-one comparison with the predicted trees, in which parent node labels are automatically derived from their daughter nodes (see Section \ref{sec_tree_eval_sub_postprocessing}).

\subsubsection{Self-Attachment}
\label{sec_tree_eval_sub_annotation_sub_self}

Aside from attaching two pieces to each other, sewing instructions often require attaching a piece to \textit{itself}\textemdash for example, a sleeve can be assembled by rolling up a rectangular piece of fabric, and sewing the two connected edges together\textemdash including repeated self-attachment in more complex garments.

To account for this phenomenon, we append a subscriped, integer-valued self-attachment counter $n$ to the label of each node $\alpha$ in each assembly tree $t$: $\alpha_n$. In cases where the assembly requires the attachment of $\alpha_n$ to itself, we include a node $\alpha_{n+1}$ and an edge $\alpha_{n+1}\rightarrow\alpha_n$ in $t$. Note that for the sake of representational simplicity, we often omit the self-attachment counter when $n=0$ in this work. 

When two pieces are attached to one another, the parent node inherits the larger self-attachment number of its daughter nodes, resulting in subtrees of the form: $\alpha_n\leftarrow\alpha\beta_{\textit{max}(n,m)}\rightarrow\beta_m$. As a consequence, each parent node label contains only one self-attachment counter, appended to the far-right edge of the node label: for example, $\text{AB}_2\leftarrow\text{ABCD}_3\rightarrow\text{CD}_3$ and $\text{A}_1\leftarrow\text{ABC}_1\rightarrow\text{BC}_0$ are valid subtrees, while $\text{AB}_2\leftarrow\text{AB}_2\text{CD}_3\rightarrow\text{CD}_3$ and $\text{A}_1\leftarrow\text{A}_1\text{BC}_0\rightarrow\text{BC}_0$ are not.

We again impose this constraint to facilitate comparison with the predicted-instruction-derived trees (see Section \ref{sec_tree_eval_sub_postprocessing}). Although this self-attachment-inheritance procedure does result in parent node labels that forget information regarding the self-attachment levels of their daughter nodes, this does not impair evaluation, as this information is still encoded within the respective subtrees that they dominate. 

\subsection{Prediction Postprocessing}
\label{sec_tree_eval_sub_postprocessing}

For each pattern $P$ in the dataset, the model in question is prompted to generate corresponding instructions $I$: in order to facilitate the tree construction procedure described in this section, we prompt the model to combine no more than two pieces in one step in $I$. We then employ a Piece-Extraction LLM, which is prompted to extract the piece labels mentioned in each step of $I$. 

Note that the instruction-generating model often does not mention intermediate pieces in the instructions, but rather single components of those pieces: for example ``attach the pocket (A) to the front piece of the skirt (B)'', when the front piece has already been attached to the rest of the skirt in prior steps. Naively extracting the pieces mentioned in such a step would result in a subtree of the form $\text{A}\leftarrow\text{AB}\rightarrow\text{B}$, when the desired outcome is of the form $\text{A}\leftarrow\text{A}\beta\rightarrow\beta$, where $\beta$ denotes the skirt to which B is attached. 

To account for this problem, we employ a rule-based procedure that replaces each mentioned piece label B with the label $\beta$ of the larger, intermediate piece to which B is attached (if applicable). 

From the pieces extracted from each step $s_k$ of $I$, we derive a subtree $t_k$ of the form $t_k=\alpha_n\leftarrow\alpha\beta_{\textit{max}(n,m)}\rightarrow\beta_m$ if two pieces $\alpha_n$, $\beta_m$ were extracted from $s_k$, or $t_k=\alpha_{n+1}\rightarrow\alpha_n$ if one piece $\alpha_n$ is extracted. These extracted subtrees are constructed to conform to the assembly-tree constraints described in Section \ref{sec_tree_eval_sub_annotation}.

Finally, we derive a tree\footnote{$T(I)$ may not necessarily be a tree, but rather a forest: for example, when the model forgets to attach a piece to the final garment.} $T(I)$ from the disjoint union of the extracted subtrees $t_k$ by gluing together nodes with the same label.

This pipeline is illustrated in Table \ref{tab:postprocessing}.

\subsection{Tree Score}
\label{sec_tree_eval_sub_score}


Let $P$ be some pattern in the dataset, with a set of associated gold trees $G(P)$ (see Section \ref{sec_tree_eval_sub_annotation}). For a given model-generated instruction $I$ for $P$, we compute its \textit{tree score} with respect to $P$ ($S_P(I)$) as the maximum $F_1$ score between $T(I)$ and each of the gold trees $g\in G(P)$ (Equation \ref{eq:1}). As each $g\in G(P)$ is represents a valid assembly order for $P$, we return the maximum $F_1$ score in order to account for the multitude of possible valid assembly orders for a given pattern.

\begin{equation}
    S_P(I)=\underset{g\in G(P)}{\textit{max}}\hspace{1mm}F_1(T(I),g)
    \label{eq:1}
\end{equation}

We compute the $F_1$ score with respect to subtrees: for each non-leaf node $\alpha\in T(I)$ and $\beta\in g$, we extract the subtrees $T(I)_\alpha$, $g_\beta$ spanning $\alpha$/$\beta$ and their immediate respective daughter nodes. We then compute the exact-match $F_1$ score between $\{T(I)_\alpha\}_{\alpha\in\textit{parents}(T(I))}$ and $\{g_\beta\}_{\beta\in\textit{parents}(g)}$ to yield $F_1(T(I),g)$.


\begin{table*}[t]
    \centering
    \scalebox{0.95}{\begin{tabular}{l|c|c}
        Instructions & Extracted Pieces & Extracted Subtree \\
        \hline
        \hline
        1. Align the Over Skirt (A) and Under Skirt (B) at the side seams. & 1: [A, B] & $\textit{AB} \rightarrow \textit{A}\hspace{2.5mm}\textit{B}$ \\
        Sew the left side seam of the Over Skirt (A) to the left side seam & & \\
        of the Under Skirt (B). & & \\
        \hline
        2. Sew the right side seam of the Over Skirt (A) to the right side & 2: [A, B] & $\textit{AB}_1 \rightarrow \textit{AB}$ \\
        seam of the Under Skirt (B). & & \\
        \hline
        3. Align the Waistband (C) with the top edge of the joined Over & 3: [C, A, B] & $\textit{S} \rightarrow \textit{AB}_1\hspace{2.5mm}\textit{C}$ \\
        Skirt (A) and Under Skirt (B). Sew the Waistband (C) to the top & & \\
        edge, ensuring the seams are evenly distributed. & & \\
        \hline
        4. Fold the Waistband (C) over to the inside of the skirt, enclosing & 4: [] & \\
        the raw edge. Stitch in place to secure. &  & \\
        \hline
        5. Hem the bottom edge of the Over Skirt (A) and Under Skirt & 5: [] & \\
        (B) to the desired length. & & \\
    \end{tabular}}
    \caption{Subtree extraction, side-by-side comparison.}
    \label{tab:postprocessing}
\end{table*}

\section{Experiments}
\label{sec_experiments}

We evaluated our proposed tree-based evaluation metric against three existing metrics\textemdash BLEU, ROUGE-L, and BERT-Score\textemdash on a set of sewing instructions generated from a variety of prompting strategies (Section \ref{sec_models}). Specifically, we compared our metric to the existing approaches with respect to correlation with error count (Section \ref{sec_experiments_sub_exp1}) and sensitivity to random permutations (Section \ref{sec_experiments_sub_exp2}).

\subsection{Data Generation}
\label{sec_models}


We employed GPT-4V (model = gpt4o, seed = 1, temperature = 0.0, max\_tokens = 3000) to generate step-by-step instructions given an image overview and a description of each of the 22 patterns in our dataset (see e.g.\hspace{1mm}Figure \ref{fig:overviewanddescrMDF359}), and as the Piece Extraction model for our evaluation metric (see Section \ref{sec_tree_eval_sub_postprocessing}).
To assess the robustness of our evaluation metric, we evaluated four prompting strategies:


\begin{itemize}
    \item \textbf{Baseline:} A zero-shot prompted model. The model is instructed to explicitly mention all connecting seams\textemdash including those that occur twice on mirrored pieces (e.g.\hspace{1mm}sleeves)\textemdash in order to accurately reflect the assembly process. As required by our evaluation metric, the model is instructed to include at most one connecting seam in each step. For the same reason, we require that the instructions explicitly mention the labels of each piece that is included in a given step. 

    \item \textbf{Few-Shot:} A few-shot prompted model that is presented with three example instructions. 

    \item \textbf{Generalized Instructions:} A model that is few-shot prompted with archetypical examples of instructions of each of the five garment types (without images), to serve as a guide for instruction generation. 

    \item \textbf{Intermediate Representations:} A model that is prompted after each generated step to generate an intermediate representation of the current state of the garment assembly process: the used pieces, the current intermediate garment(s), and the unused pieces in the assembly process.
\end{itemize}

Example prompts for these strategies are located in Figures \ref{fig:baselineprompt}-\ref{fig:intermrepprompt} in the Appendix.

\paragraph{Instruction Generation.}

We evaluated all possible combinations of these prompting strategies: few-shot prompting, few-shot prompting with generalized instructions, etc. Including the baseline approach, this results in eight possible combinations across the 22 patterns in our dataset, yielding 176 generated sets of instructions. 

To enable an accurate, one-to-one comparison to BLEU, ROUGE-L, and BERT-Score, we hand-crafted gold instructions for each of the patterns in our dataset, that align with the style and constraints imposed by the tree-based evaluation metric: i.e.\hspace{1mm}always mentioning piece labels, connecting at most one seam per step, etc.

\subsection{Error Reflection}
\label{sec_experiments_sub_exp1}

We first compare our metric to existing approaches with respect to their capacity to reflect errors of spatiotemporal reasoning in generated instructions. 

\subsubsection{Experimental Setup}
\label{sec_experiments_sub_exp1_sub_setup}

For each of the 176 model-generated instructions, we computed BLEU, ROUGE-L, and BERT-Score with respect to the gold text, and computed our tree-score between the extracted tree and the annotated gold trees (see Section \ref{sec_tree_eval}). 

We manually annotated the model-generated instructions for errors, counting errors across four categories: (i) incorrect assembly operations, (ii) missing assembly operations, (iii) incorrect order of assembly operations, and (iv) conflicting information/incorrect use of terminology. We then computed the correlation between error count and score for each of the four evaluation metrics. 

\begin{table}[t]
\centering
\begin{tabular}{l|ll}
Metric & Corr. & $p$ \\
 \hline
 \hline
 BLEU & -0.111 & .1432\\
 ROUGE-L & -0.179 & < .05\\
 BERT-Score & -0.080 & .2915\\
 Tree Score (ours) & \textbf{-0.599} & \textbf{< .001}\\
\end{tabular}
\caption{Pearson's correlation of evaluation metrics with the number of errors per number of steps (df = 174).}
\label{tab:corrtable}
\end{table}

\subsubsection{Results}
\label{sec_experiments_sub_exp1_sub_results}

There is a significant, moderate negative correlation between tree score and number of errors (see Table \ref{tab:corrtable})\textemdash substantially lower than that observed for BLEU, ROUGE-L, and BERT-Score. In fact, BLEU and BERT-Score are not significantly correlated with error rate at all: these metrics are entirely unable to detect such errors.

These results clearly demonstrate our evaluation metric's superior ability to reflect errors of spatiotemporal reasoning. 

\subsection{Robustness to Permutation}
\label{sec_experiments_sub_exp2}


Next, we evaluated the robustness of the four evaluation metrics with respect to artificially-constructed counterfactual examples that are designed to disrupt spatiotemporal correctness. 

\subsubsection{Experimental Setup}
\label{sec_experiments_sub_exp2_sub_setup}

In this experiment, we randomly permuted the steps of each of the 22 instructions generated by the baseline prompting approach (see Section \ref{sec_models}): this constructs a set of instructions that is stylistically and lexically similar to the gold instructions, but entirely unexecutable due to the nonsensical ordering of the instruction steps. 

We then computed the BLEU, ROUGE-L, BERT-Score, and tree score for each of the permuted instructions, and compared these scores to each of the metrics respective scores for the original, unpermuted baseline-approach instructions. 

\begin{table}
\centering
\scalebox{0.95}{\begin{tabular}{l|lll}
Metric & Baseline & Shuffled & $\Delta$ \\
\hline
\hline
BLEU & 0.127 & 0.126 & -0.001 \\
ROUGE-L & 0.377 & 0.322 & -0.055 \\
BERT-Score & 0.881 & 0.878 & -0.003 \\
Tree Score (ours) & 0.512 & 0.272 & \textbf{-0.240} \\
\end{tabular}}
\caption{Mean evaluation scores across the baseline and randomly-permuted instructions.}
\label{table_permuted}
\end{table}

\subsubsection{Results}
\label{sec_experiments_sub_exp2_sub_results}

The results of this experiment (Table \ref{table_permuted}) clearly demonstrate that our metric is more sensitive to errors of spatiotemporal reasoning than existing evaluation metrics. The average tree score drops by almost 50\% from the baseline to the permuted examples, reflecting the decrease in correctness of the artificially-constructed instructions.

In comparison, ROUGE-L score decreases by only 15\%, while BLEU and BERT-Score fail to decrease by any meaningful amount: these metrics entirely fail to capture any difference in terms of correctness between the actual model-generated instructions and the randomly-permuted instructions.

\subsection{Alignment with Human Judgment}

To evaluate the degree to which our tree score metric aligns with human judgment, we recruited 20 participants with experience in sewing garments to rate the model-generated instructions of Section \ref{sec_models}. All human evaluations were conducted on the LingoTurk platform \cite{pusse-etal-2016-lingoturk}.


\subsubsection{Experimental Setup}



\paragraph{Data}

Each participant was asked to evaluate 4 sets of sewing instructions since a brief pilot study showed that 4 sets can be completed in about 30 minutes.
We collected ratings 110 sets of instructions, a subset of the dataset that we used in previous experiments (spanning all 22 patterns).


\paragraph{Task}

We provided the participants with the overview image/description of the pattern pieces\textemdash the same context available to the generation model\textemdash and the model-generated instructions.

We employed a step-by-step evaluation: participants were presented with one step of the instructions at a time, and gave individual ratings for each step.
This method reduces the mental load on the participants and results in more detailed ratings, enabling a more fine-grained analysis.



At each step, the participants answered the following questions with a rating on a 5-point Likert Scale:

\begin{itemize}
    \item[S1.] Does this step make sense with regard to the picture?
    \item[S2.] Does this step make sense with regard to the pattern piece descriptions?
    \item[S3.] Does this step make sense with regard to the previous step?
    \item[S4.] Is this step clear enough to follow?
    \item[S5.] Is the terminology in this step used correctly (e.g. seam names)?
\end{itemize}

Once the ratings were provided, the next step was displayed on the screen\textemdash the previous steps always remained visible. An example of the participant interface is given in Figure \ref{fig:experimentmain} in the Appendix.


After rating each step, the participants were asked to give two ratings (also on a 5-point Likert Scale) of the instructions as a whole:

\begin{itemize}
    \item[I1.] Is it possible to assemble the pieces following the given instructions?
    \item[I2.] This pattern was designed to be a <garment type>. Do you think the assembly steps lead to the correct outcome?
\end{itemize}


\paragraph{Data Aggregation}

For each step-level question, we aggregated the step-by-step ratings into three instruction-level summarized ratings ($5\times3=15$ summarized ratings per instruction): the mean rating across steps, the proportion of ratings above 3, and the proportion of ratings below 3.

\begin{figure}
    \centering
    \includegraphics[width=\columnwidth]{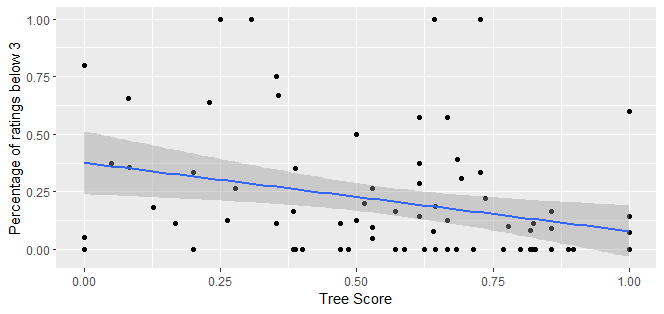}
    \caption{Proportion of below-3-ratings for S3, plotted against tree score.}
    \label{fig:below3ratings}
\end{figure}

\subsubsection{Results}

\begin{table}
    \centering
    \scalebox{0.72}{\begin{tabular}{l|lllll|ll}
     & S1 & S2 & S3 & S4 & S5 & I1 & I2 \\
    \hline
    \hline
    Tree (ours) & 0.14 & 0.12 & \textbf{0.23} & 0.15 & 0.12 & 0.05 & 0.16\\
    BLEU & \textbf{-0.26}  & \textbf{-0.29} & -0.18 & -0.19 & -0.22 & -0.18 & -0.07\\
    ROUGE-L & -0.04 & -0.12 & -0.02 & -0.02 & -0.12 & -0.11 & 0.03\\
    BERT & -0.12 & -0.20 & -0.10 & -0.08 & -0.13 & -0.14 & 0.00\\
    \end{tabular}}
    \caption{Pearson's correlation between evaluation scores, and mean step-level (left; S1-5) and instruction-level (right; I1-2) human ratings. Significant values ($p < .05$) are indicated in bold.}
    \label{tab:corrratings}
\end{table}

Mean step-level ratings from each rating category are weakly, positively correlated with tree score (Table \ref{tab:corrratings}), although only the correlation with S3 is significant ($r(78) = 0.23$, $p < .05$). In addition, there is a significant, weak negative correlation between tree score and the proportion below 3 for S3 ($r(78) = -0.29$, $p < .01$; see Figure \ref{fig:below3ratings}).

For the traditional evaluation metrics, we observe a negatively correlated relationship with human judgment (see Table \ref{tab:corrtable}) suggesting that these measures do not align with the perceived quality of the instructions.

Under the assumption that human judgment is most likely to consider spatiotemporal consistency at the step (rather than instruction) level, the correlation of S3 with tree score indicates that our metric reflects human assumptions regarding spatiotemporal consistencies to a higher degree than traditional similarity scores.

\section{Conclusion}
\label{sec_conclusion}

We introduced a novel, tree-based evaluation metric that is designed to more accurately reflect spatiotemporal correctness in generated instructions (see Section \ref{sec_tree_eval}). This metric affords the flexibility required to capture multiple orders of assembly, while reflecting the spatial and temporal aspects of reasoning required for instruction generation.

In the domain of sewing instructions, we showed that our metric better (negatively) correlates with error counts than existing evaluation metrics (see Section \ref{sec_experiments}), demonstrating this metrics superior ability to reflect \textit{correctness}\textemdash rather than simple textual similarity. Further experiments show that our metric is far more robust to artificially-constructed instructions specifically designed to confound metrics that rely purely on similarity. 

In addition, the results of a human evaluation study indicate that human perception of coherence in step-by-step sewing instructions is positively correlated with our proposed evaluation metric, but not with traditional similarity measures.

These results indicate that our tree score is a more meaningful metric for tasks such as sewing-instruction generation, in which step-by-step correctness and consistency is far more important than stylistic and lexical resemblance.




\section*{Limitations}

\paragraph{Reflection of Attachment Method.}
In order to represent the exactly how the pattern pieces have to be connected to form the finished garment, it is important to know which edges of which pieces connect to which other edges on which other pieces.
The approach that is introduced in this work only represents which pieces are attached to which other pieces and when.

However, the same applies to the textual sewing instructions: the gold instructions as well as the generated instructions assume that the reader has background knowledge in sewing and do not explicitly mention the edges of the pieces in many cases.
This makes it impossible to reliably extract information about the edges of the pieces as this information might not be present in the textual instructions to begin with.

\paragraph{Dependence of the Evaluation Metric on Input Format.}
Further, the tree-based evaluation only works as intended when the generated instructions follow specific constraints that allow the tree extraction algorithm to function.
This naturally limits the number of scenarios where our tree-based evaluation approach is practical.

\paragraph{Garment Complexity.}
In addition, our tree-based evaluation is limited to the evaluation of simple garments.
In order to be able to evaluate more complex garments, the tree-based evaluation framework must be expanded (for example to accommodate patterns that include multiple left and right copies of the same piece).
We leave the investigation of the extension of our metric to accommodate more complex sewing patterns\textemdash as well as other instruction generation tasks\textemdash to future work.

\paragraph{Self-Attachment.}
Another limitation concerning the tree-based evaluation is that the property of an intermediate product is always represented by the same node, regardless of how it was assembled: this is true for most cases (most shirts, skirts, dresses) but not all (see Appendix \ref{sec_appendix}).

\paragraph{Node Naming.}
Errors propagating up through the tree builds a further limitation, as the mistakes that are made earlier in the assembly process affect the tree more heavily. This is due to the fact that these trees are constructed
bottom-up, and the names of the later nodes are defined by the previous assembly steps.

Although it makes sense to punish early mistakes more heavily\textemdash because later steps cannot be followed when the earlier steps produced an incorrect intermediate garment\textemdash the current tree-based evaluation metric assigns a similar score to instructions with frequent early mistakes, regardless of whether the following steps are correct.

\paragraph{Domain}
We show that our proposed evaluation approach outperforms traditional evaluation metrics in the presented setting, which however is restricted to the very specific task of sewing instruction generation. 
While we believe this CFG-based tree evaluation framework to be a generalizable approach that could be translated to other tasks where textual instructions need to be evaluated with respect to real world 3D applications, such as furniture building, puzzle games or Minecraft, the ability to generalize our approach to these domains remains to be explored in future work.

Conversely, as our CFG-based evaluation metric delivers a structured representation of a physical assembly strategy, future work could investigate how training on these representation might improve the performance of language models in tasks requiring complex spatiotemporal understanding.

\bibliography{anthology,custom}

\appendix

\section{Appendix}
\label{sec_appendix}
\subsection{Cutting on the fold}
\begin{figure}
    \centering
    \includegraphics[width=\linewidth]{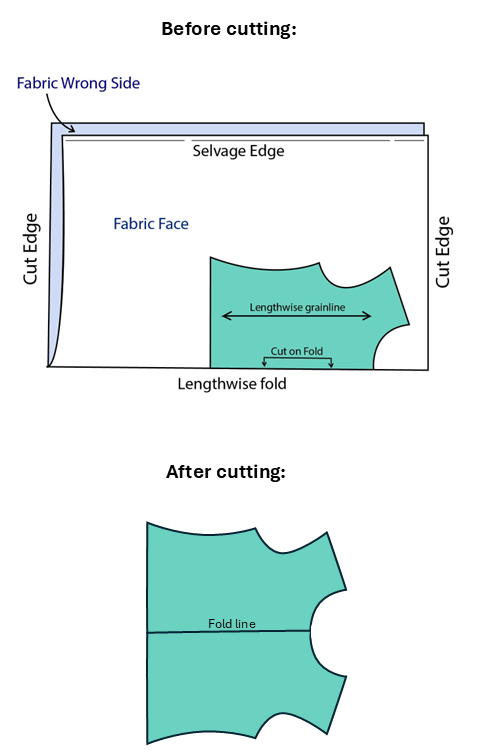}
    \caption{Illustration of cutting a pattern piece ``on the fold'' (source: \url{https://i.pinimg.com/originals/61/d9/2f/61d92fede2df0c2a159caac9d96bbafb.jpg}).}
    \label{fig:cuttingonthefold}
\end{figure}

Figure \ref{fig:cuttingonthefold} illustrates the principle of cutting pattern pieces on the fold of the fabric. 
The paper pattern only shows half of the piece and is placed on the folded edge of the fabric before cutting.
Like this, two layers of fabric are cut at the same time to reveal the full piece once the fabric is unfolded again.
This is commonly done to easily obtain a symmetrical piece (here a front bodice).

\subsection{Human Evaluation Experiment}
\begin{figure*}
    \centering
    \includegraphics[width=\textwidth]{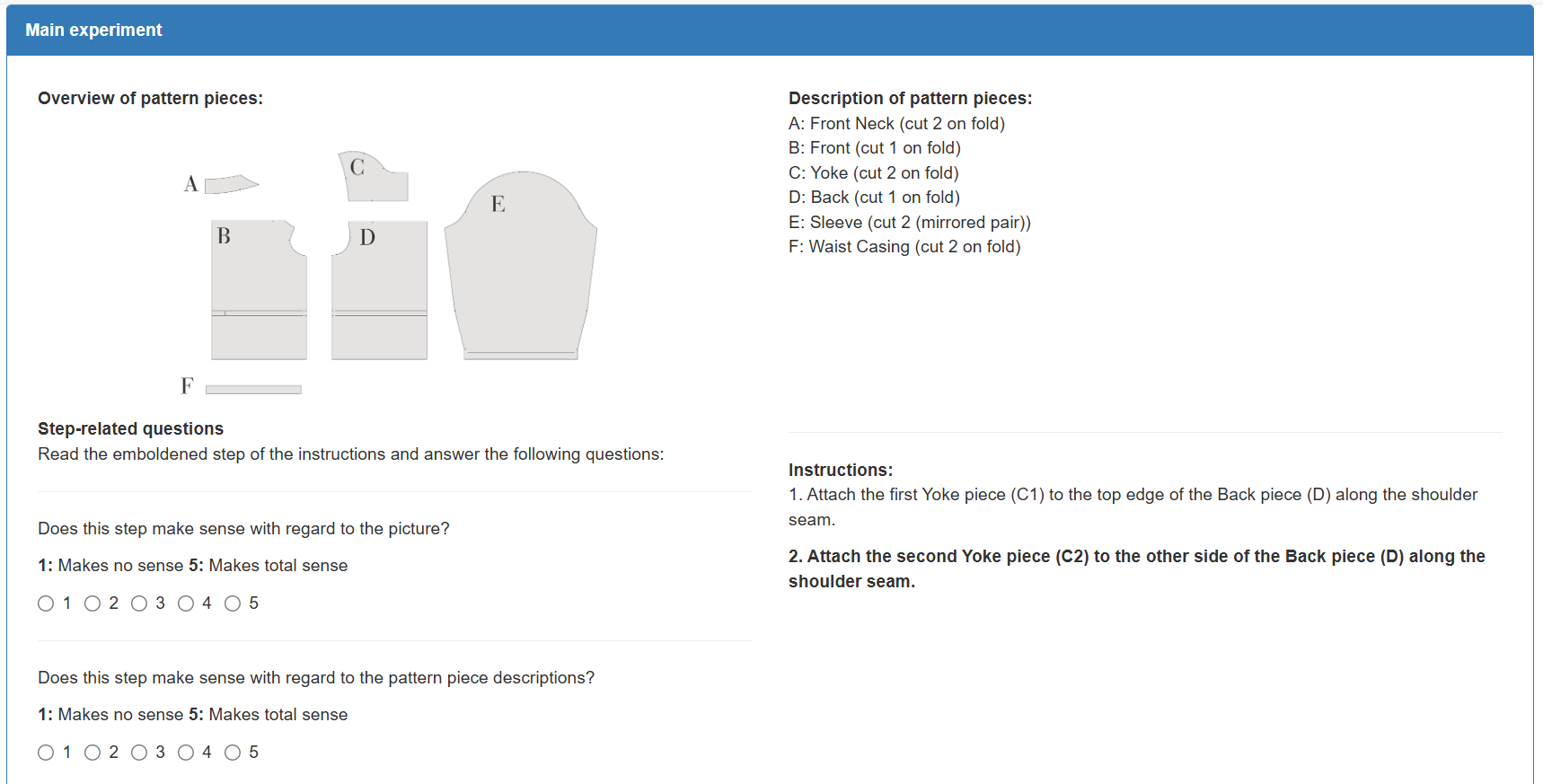}
    \caption{Example of the human-evaluation experiment participant interface.}
    \label{fig:experimentmain}
\end{figure*}

Figure \ref{fig:experimentmain} shows the experiment interface with the overview of the pattern pieces in the top left corner, the description of the pattern pieces in the top right corner, the step-by-step instructions in the bottom right corner and the evaluation questions in the bottom left corner.

\subsection{Self-Attachment Limitation}

\begin{figure}[h!]
    \centering
    \includegraphics[width=\columnwidth]{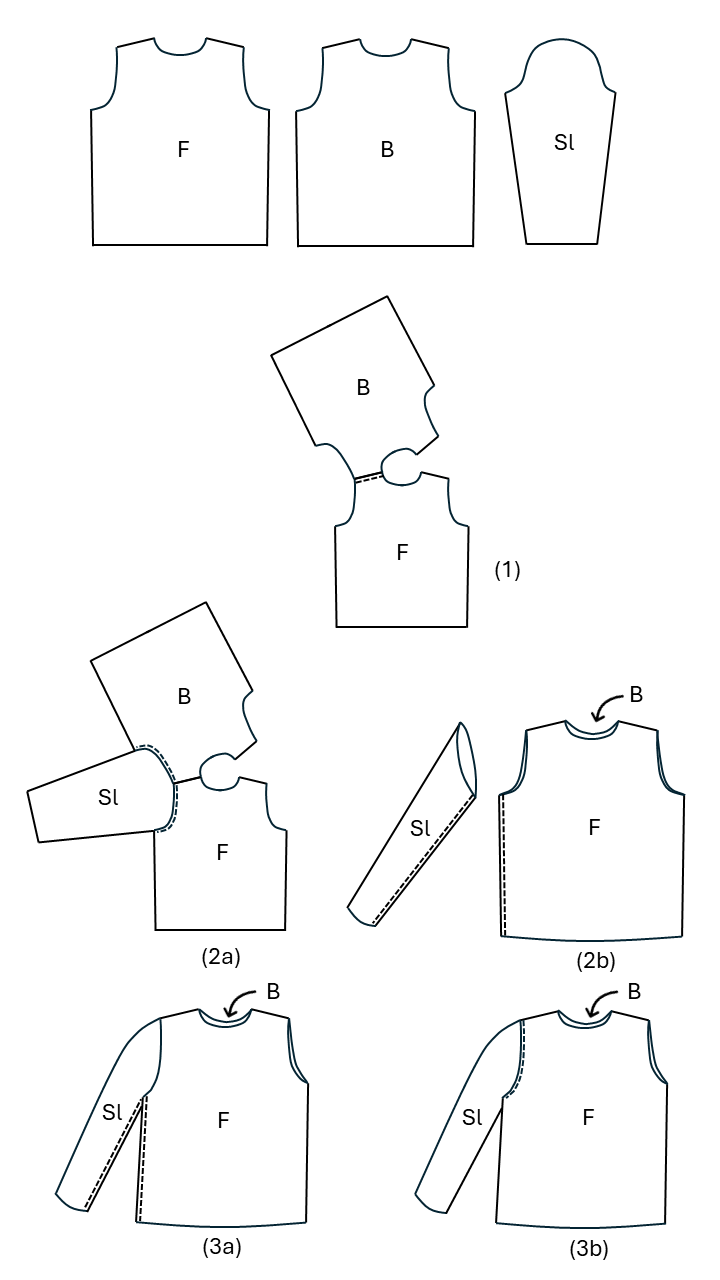}
    \caption{Two ways of assembling a shirt (a and b): the pieces F (front), B (back), and Sl (left sleeve) can be sewn together following either sequence of steps: $(1)\rightarrow(2a)\rightarrow(3a)$ or 
    $(1)\rightarrow(2b)\rightarrow(3b)$.}
    \label{fig:shirtassembly}
\end{figure}

\begin{figure}
    \centering
    \includegraphics[width=\columnwidth]{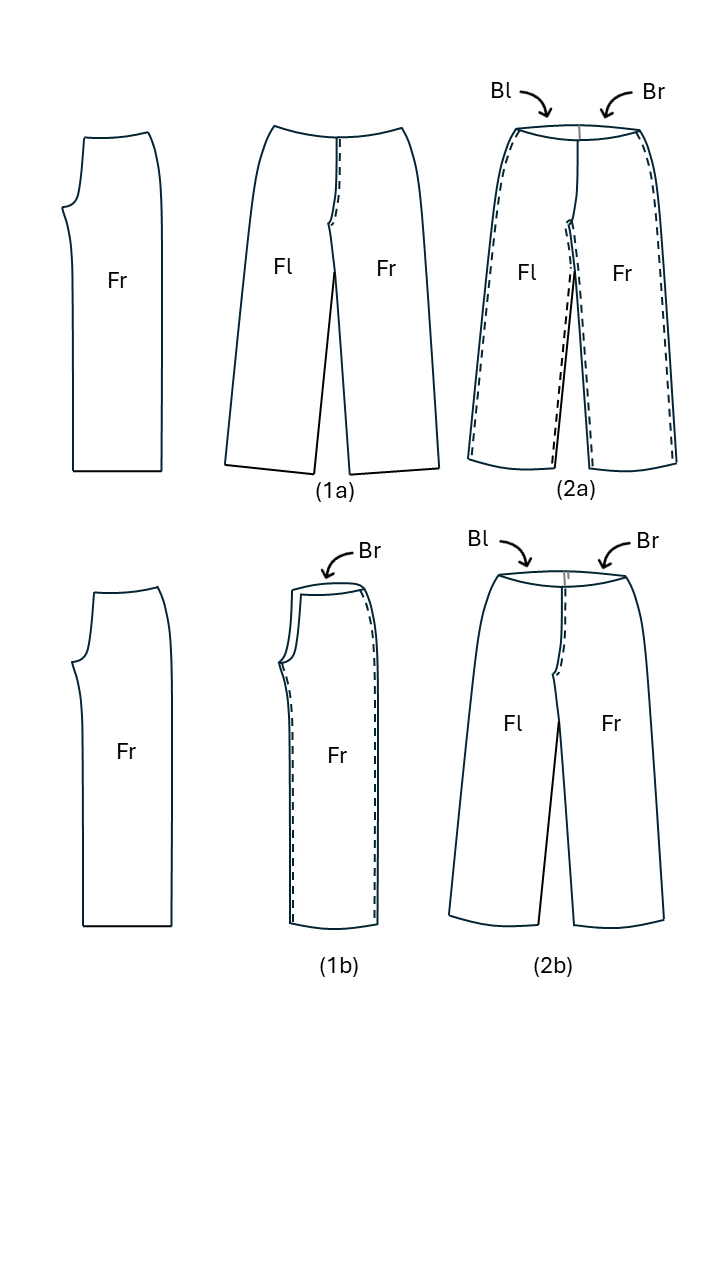}
    \caption{Two ways of assembling a pair of pants (a and b), Fl: left front piece, Fr: right front piece, Bl: left back piece, Br: right back piece, dotted lines indicate where seams are sewn}
    \label{fig:pantsassembly}
\end{figure}

As an example, Figure \ref{fig:pantsassembly} shows two strategies for assembling a pair of pants.
A pair of pants usually consists of 4 pieces, two front pieces (here Fl and Fr) and two back pieces (here Bl and Br).
The picture shows the right front piece Fr.
The right back piece Br looks the same, while the left pieces Fl and Bl are mirrored versions of Fr.

With those 4 pieces, we have the opportunity to either attach the left and right pieces to each other at the crotch seam forming one front piece (FlFr) and one back piece (BlBr) (see (1a)) or we attach the two left pieces to each other (BlFl) and the two right pieces (BrFr).

The first approach ($(1a) \rightarrow (2a)$) needs 3 more seams: the left side seam, the right side seam and the inseam (reaching from the bottom of one leg all the way to the bottom of the other leg; see Figure \ref{fig:pantsassembly}).
These 3 seams can be executed in any order, leading to the following rules, where the first seam attaches the front and back pieces to each other while the second and third seams are self-attachment steps:

\begin{subequations}
\begin{align}
&\text{BlBrFlFr} \rightarrow \text{BlBr}\hspace{2.5mm}\text{FlFr} \\
&\text{BlBrFlFr}_1 \rightarrow \text{BlBrFlFr} \\
&\text{BlBrFlFr}_2 \rightarrow \text{BlBrFlFr}_1
\end{align}
\end{subequations}

The second approach forms the left and right pieces into two tubes which are attached along the crotch seam in the next step to form the final pair of pants (see Figure \ref{fig:pantsassembly}).
Adhering to the established constraints, this results in the following rules:

\begin{subequations}
\begin{align}
&\text{BlFl}_1 \rightarrow \text{BlFl} \\
&\text{BrFr}_1 \rightarrow \text{BrFr} \\
&\text{BlBrFlFr}_1 \rightarrow \text{BlFl}_1\hspace{2.5mm}\text{BrFr}_1    
\end{align}
\end{subequations}

We can see that even though the number of seams and finished product are the same when comparing the two assembly strategies, the representation of the finished product is different.

Changing the constraints to 
$parentnode\_x \rightarrow child1\_y child2\_z$ where x = y+z, conversely, would break the tree logic for the shirt example in Figure \ref{fig:shirtassembly}.

\begin{figure*}[h!]
    \centering
    \includegraphics[width=\textwidth]{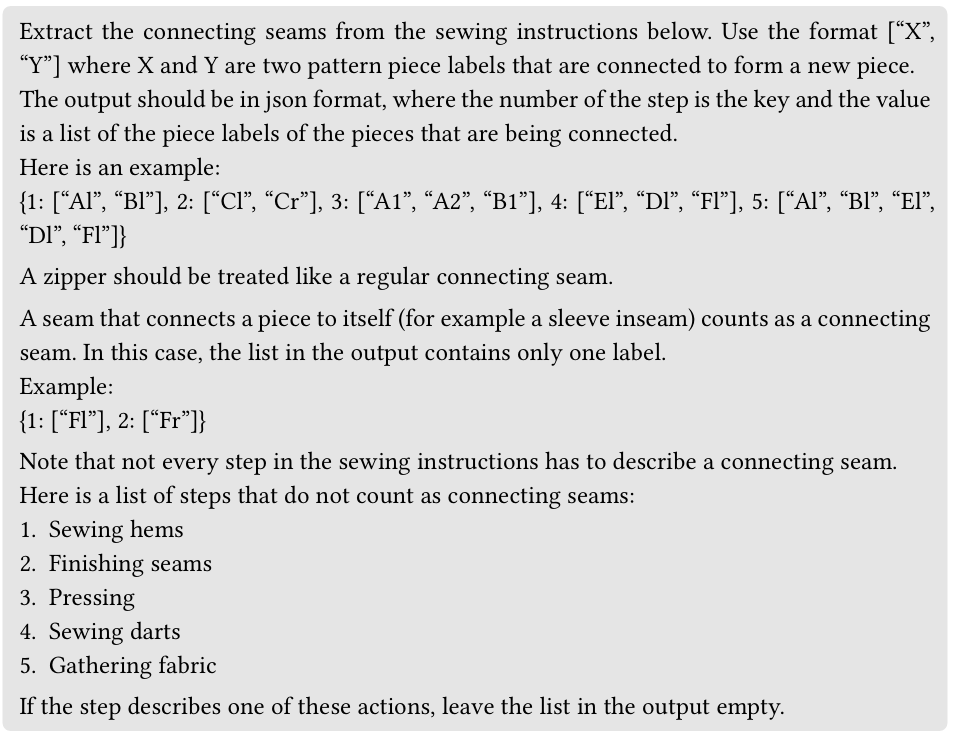}
    \caption{Piece Extraction Prompt}
    \label{fig:extractionprompt}
\end{figure*}

\subsection{Instruction Generation Prompts}
\label{sec_a_baselineprompt}

\begin{figure*}[h!]
    \centering
    \includegraphics[width=\textwidth]{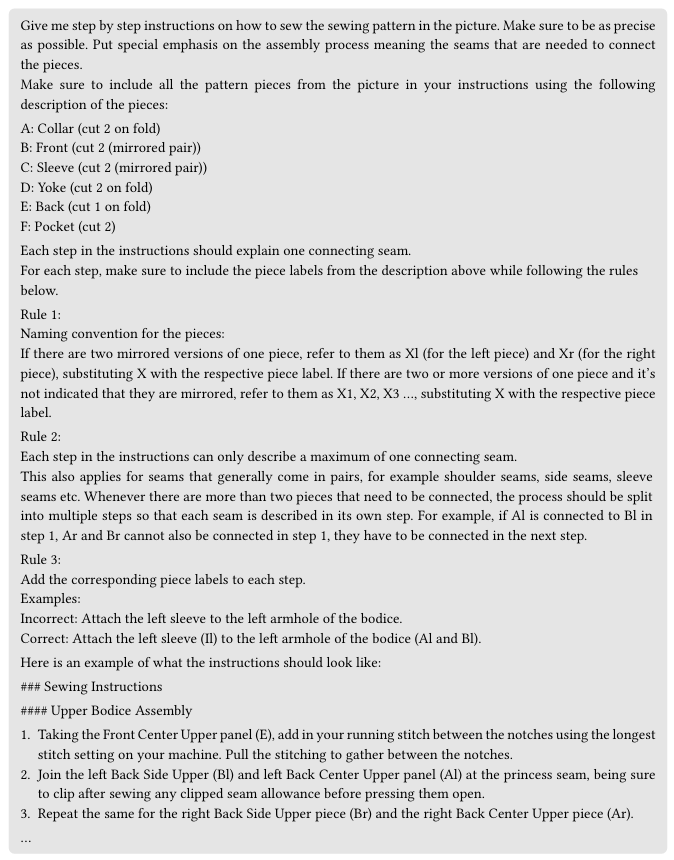}
    \caption{Baseline Prompt}
    \label{fig:baselineprompt}
\end{figure*}


\begin{figure*}[h!]
    \centering
    \includegraphics[width=\textwidth]{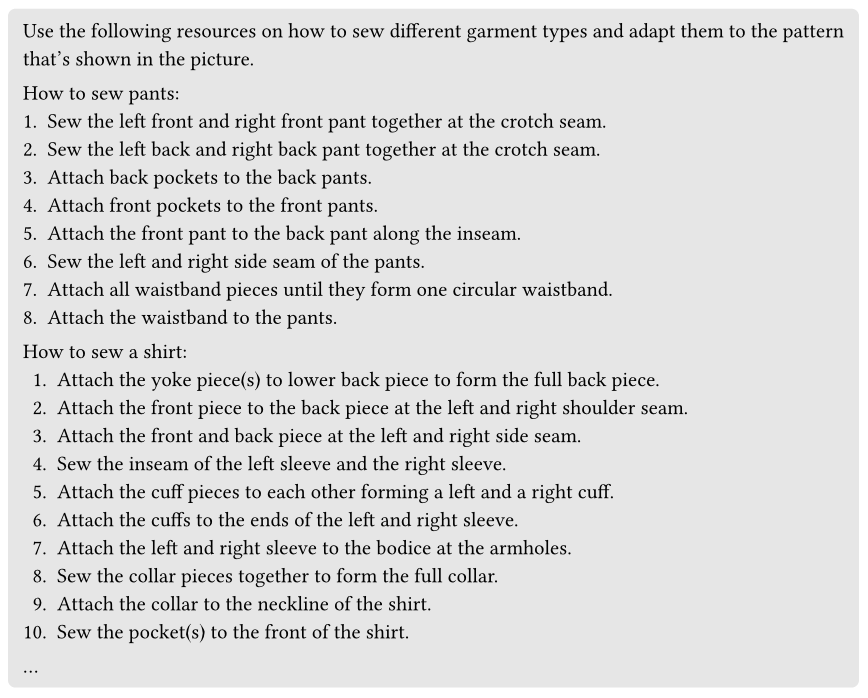}
    \caption{Generalized Instructions Prompt}
    \label{fig:geninstrprompt}
\end{figure*}

\begin{figure*}[h!]
    \centering
    \includegraphics[width=\textwidth]{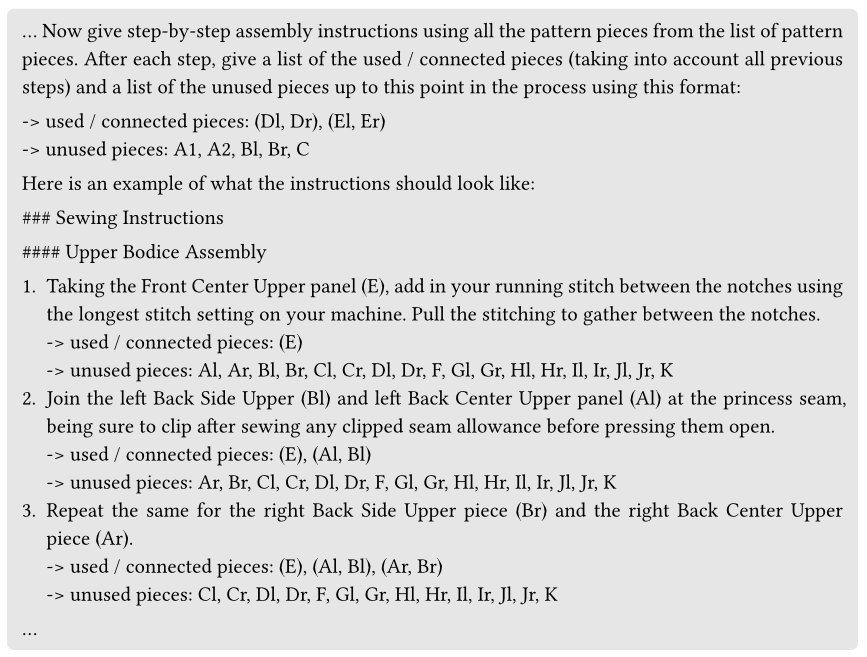}
    \caption{Intermediate Representation Prompt}
    \label{fig:intermrepprompt}
\end{figure*}


\end{document}